\documentclass[sigconf,screen,authorversion]{acmart}
\usepackage{graphicx} % Required for inserting images
\usepackage{subcaption}
\usepackage{multirow}

\title[Implementing and Benchmarking LCA on Loihi 2]{Implementing and Benchmarking the Locally Competitive Algorithm on the Loihi 2 Neuromorphic Processor}

\author{Gavin Parpart}
\email{gavin.parpart@pnnl.gov}
\orcid{0000-0003-4151-369X}
\affiliation{%
  \institution{Pacific Northwest National Laboratory}
  \streetaddress{902 Battelle Blvd}
  \city{Richland}
  \state{WA}
  \country{USA}
  \postcode{99354}
}

\author{Sumedh R. Risbud}
\email{sumedh.risbud@intel.com}
\orcid{0000-0003-4777-1139}
\affiliation{
    \institution{Intel Labs}
    \department{Neuromorphic Computing Lab}
    \city{Santa Clara}
    \state{CA}
    \country{USA}
    \postcode{95054}
}

\author{Garrett T. Kenyon}
\email{gkenyon@lanl.gov}
\orcid{0000-0003-4836-3938}
\affiliation{%
  \institution{Los Alamos National Laboratory}
  \streetaddress{P.O. Box 1663, MS-B256}
  \city{Los Alamos}
  \state{NM}
  \country{USA}
  \postcode{87545}
}

\author{Yijing Watkins}
\orcid{0000-0003-3094-2421}
\email{yijing.watkins@pnnl.gov}
\orcid{0000-0003-1748-354X}
\affiliation{%
  \institution{Pacific Northwest National Laboratory}
  \streetaddress{902 Battelle Blvd}
  \city{Richland}
  \state{WA}
  \country{USA}
  \postcode{99354}
}

\renewcommand{\shortauthors}{Parpart, et al.}

\ccsdesc[500]{Hardware~Neural systems}
\ccsdesc[500]{Hardware~Power and energy}

\date{April 2023}

\copyrightyear{2023} 
\acmYear{2023} 
\acmConference[ICONS '23]{International Conference on Neuromorphic Systems}{August 1--3, 2023}{Santa Fe, NM, USA}
\acmBooktitle{International Conference on Neuromorphic Systems (ICONS '23), August 1--3, 2023, Santa Fe, NM, USA}
\acmPrice{15.00}
\acmDOI{10.1145/3589737.3605973}
\acmISBN{979-8-4007-0175-7/23/08}

\begin{document}

\begin{abstract}
    Neuromorphic processors have garnered considerable interest in recent years for their potential in energy-efficient and high-speed computing. The Locally Competitive Algorithm (LCA) has been utilized for power efficient sparse coding on neuromorphic processors, including the first Loihi processor \cite{appletospikes, loihi1}. With the Loihi 2 processor enabling custom neuron models and graded spike communication, more complex implementations of LCA are possible \cite{loihi2}. We present a new implementation of LCA designed for the Loihi 2 processor and perform an initial set of benchmarks comparing it to LCA on CPU and GPU devices. In these experiments LCA on Loihi 2 is orders of magnitude more efficient and faster for large sparsity penalties, while maintaining similar reconstruction quality. We find this performance improvement increases as the LCA parameters are tuned towards greater representation sparsity.
    Our study highlights the potential of neuromorphic processors, particularly Loihi 2, in enabling intelligent, autonomous, real-time processing on small robots, satellites where there are strict SWaP (small, lightweight, and low power) requirements. By demonstrating the superior performance of LCA on Loihi 2 compared to conventional computing device, our study suggests that Loihi 2 could be a valuable tool in advancing these types of applications. Overall, our study highlights the potential of neuromorphic processors for efficient and accurate data processing on resource-constrained devices.
\end{abstract}

\maketitle

\section{Introduction}

In recent years, the concept of neuromorphic computing has gained popularity in the field of artificial intelligence and machine learning. By emulating the structure and function of biological neural systems, this computing paradigm provides power-efficient and high-speed computing capabilities. Unlike traditional von Neumann computers, neuromorphic computing stores and processes information locally within the same unit, which eliminates the need to move data around. This reduced communication bandwidth allows the chip to perform operations in a highly energy efficient manner compared to traditional computing architectures, consuming only the minimum amount of energy required to perform a given computation. Additionally,  the fine grained parallelism allows very large numbers of neurons to work simultaneously, enabling faster processing of large amounts of data. One example of neuromorphic hardware is the Loihi 2 \cite{loihi2} neuromorphic processor, which mimics the structure and function of the human brain, enabling efficient and parallel processing of data. Due to it's intrinsic speed and energy efficiency, neuromorphic computing has the potential to revolutionize various computer vision applications, including efficient and robust classification and object detection.

Sparse coding is a key technique with applications in neuromorphic computing. It models the behavior of V1 simple cell receptive fields \cite{zhu2013visual} and can acquire features in an unsupervised scenario for machine learning applications. Sparse coding algorithms use an over-complete set of non-orthogonal basis functions, known as feature vectors, to find a sparse combination of non-zero activation coefficients that can most accurately reconstruct each input image. The Locally Competitive Algorithm (LCA) \cite{rozell2008sparse} is a biologically plausible implementation of sparse coding. LCA has primarily been developed for computer vision, with successful applications in denoising \cite{Carroll_denoising}, up-sampling \cite{LCA_event_recons}, compression \cite{LCA_compression}, and image classification \cite{dictionarylearning,TetiKMM22}. Furthermore, LCA is highly compatible with neuromorphic computing as it maps the feature vectors to neurons which compete to sparsely reconstruct the input.

Despite the potential of LCA on neuromorphic computing systems, there is a lack of benchmark studies that compare the performance of different hardware platforms executing LCA. In \cite{loihi1}, LCA on Loihi 1 is compared against a CPU, showing strong performance and efficiency improvements as the problem size increases. However, the algorithm design was substantially modified from \cite{rozell2008sparse} and other platforms like GPU and edge devices aren't considered. To address this gap, we implement fixed-sized 1-layer and 2-layer LCA with varying V1 and residual thresholds on different hardware platforms, including the Loihi 2 neuromorphic processor, A100 GPU, M1 CPU, and Jetson Nano edge GPU. We measure their performance based on reconstruction error, throughput, and dynamic energy consumption. Our results show that at large sparsity penalties the Loihi 2 neuromorphic processor outperforms conventional computing devices in terms of throughput and dynamic energy consumption while maintaining the same sparsity and reconstruction quality. This study suggests that Loihi 2 is a viable option for enabling intelligent, autonomous, real-time processing on small robots, drones, and satellites with small, lightweight, and low-power requirements. Overall, this study provides valuable insights into the performance of different hardware platforms in executing LCA and highlights the potential of neuromorphic computing for efficient and performant computer vision applications.

% Loihi 2 \cite{loihi2}

% LCA has previously been implemented on the first Loihi neuromorphic processor \cite{appletospikes, loihi1}.

\subsection{Locally Competitive Algorithm (LCA)}
The Locally Competitive Algorithm (LCA) seeks to find a sparse representation $\mathbf{a}$ of an input vector $\mathbf{X}$ in terms of dictionary elements $\phi_i \in \Phi$. The objective of minimizing reconstruction error $\mathbf{X} - \Phi \textbf{a}$ and sparsity is formalized by the following energy function:
\begin{equation}
E(\mathbf{X},\Phi, \mathbf{a})=\min\limits_{ \{\mathbf{a}, \, \Phi \} } \left[  \, \frac{1}{2}  ||  \mathbf{X} - \Phi  \mathbf{a} ||^2 +	\lambda || \mathbf{a} ||_1\right],
\label{eq:SC}
\end{equation}

LCA finds a local minimum of the cost function defined in Eq.~(\ref{eq:SC}) by introducing the dynamical variables (membrane potentials) $\mathbf{u}$ such that the output~$\mathbf{a}$ is given by a soft-threshold transfer function, whose threshold is given by the sparsity tradeoff parameter~$\lambda$ \cite{rozell2008sparse}:

\begin{equation}
     \mathbf{a} = T_\lambda(\mathbf{u}) = 
    \begin{cases} 
      \mathbf{u} -  \lambda,  & \mathbf{u} > \lambda \\
      0,  & \textrm{otherwise}
   \end{cases}
\end{equation}

The cost function defined in Eq.~(\ref{eq:SC}) is then minimized by taking the gradient of the cost function with respect to $\mathbf{a}$ and solving the resulting set of coupled differential equations for the membrane potentials $\mathbf{u}$: 

\[
\dot{\mathbf{u}} \propto -\dfrac{\partial E}{\partial \mathbf{a}} = -\mathbf{u} + \mathbf{\Phi}^T \mathbf{X} - T_\lambda(\mathbf{u})\{\mathbf{\Phi}^T \mathbf{\Phi} -1 \}
\]

\begin{equation}
 = -\mathbf{u} + \mathbf{\Phi}^T \{\mathbf{X} - \mathbf{\Phi}T_\lambda(\mathbf{u})\} +  T_\lambda(\mathbf{u})
\label{eq:SC1}
\end{equation}

\subsection{One and Two Layer LCA}

The neurons described in Eq.~(\ref{eq:SC1}) can be structured as either a one or two layer model. 

\begin{figure}
\centering
\begin{subfigure}[figtopcap]{0.23\textwidth}
\centering
\includegraphics[width=0.99\linewidth]{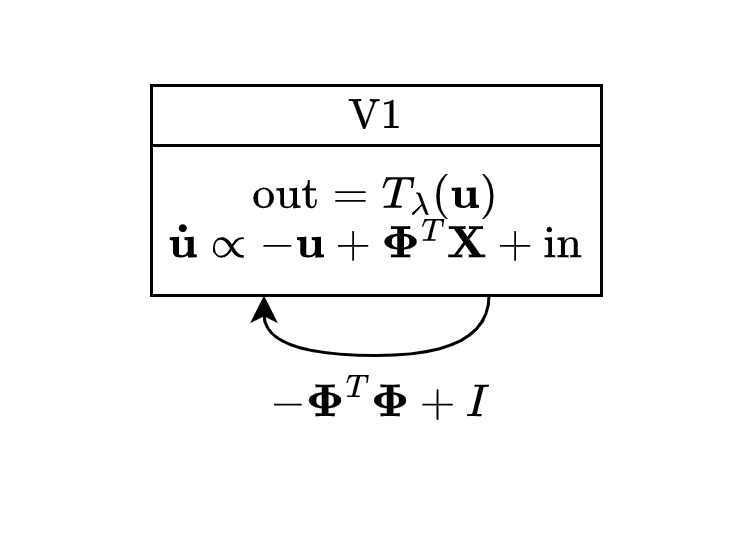}
\caption{One Layer}
\label{fig:1_layer}
\end{subfigure}%
\begin{subfigure}[figtopcap]{0.23\textwidth}
\centering
\includegraphics[width=0.99\linewidth]{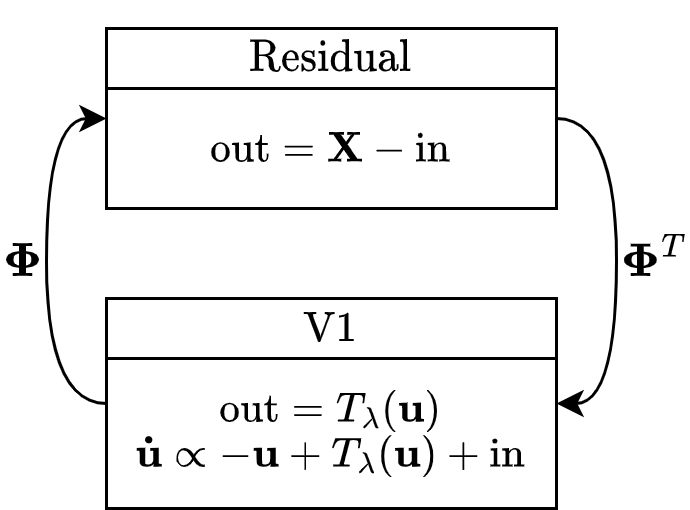}
\caption{Two Layer}
\label{fig:2_layer}
\end{subfigure}
\caption{Structure of One and Two Layer LCA Models}
\Description{TODO describe figure}
\label{fig:models}
\end{figure}

In the 1-layer model, the dynamics are structured as a single recurrent layer (Figure \ref{fig:1_layer}). We refer to these neurons as V1 as they model the behavior of V1 simple cells. There is a V1 neuron for every element of the dictionary. On traditional hardware platforms, this is a more efficient implementation than 2-layer. 
% TODO: check consistency of 1-layer and 2-layer spelling

In the 2-layer model, the reconstruction is separated into its own layer (Figure \ref{fig:2_layer}). In this residual layer, there is a neuron for every element of the input $\mathbf{X}$. This structure allows for easier implementation of convolutional LCA \cite{schultz2014replicating}, and provides a local update rule when performing dictionary learning as given in \cite{sparse_V1}.

\section{Loihi 2 Implementation}

\subsection{V1 Neurons}
The V1 neuron models directly implement the dynamics from Figure \ref{fig:models}. The voltage $\mathbf{u}$ is stored as a 24-bit signed integer. Changes to the voltage are discretized with a time-constant $\tau$:
\begin{equation}
\begin{split}
\mathbf{u}_{t+1} & = \mathbf{u}_{t} + \tau(-\mathbf{u}_{t} + b + \textrm{in} ) \\
& = \mathbf{u}_{t} (1 -\tau) + \tau~b + \tau~\textrm{in}
\end{split}
\end{equation}
where $b= \mathbf{\Phi}^T \mathbf{X}$ for one-layer LCA, corresponding to lateral competition (Figure \ref{fig:1_layer}), and $b=T_\lambda(\mathbf{u})$ for 2-layer LCA, which compensates for self-competition when the input arises from a residual error layer (Figure \ref{fig:2_layer}). In the one-layer $\tau b$ is pre-computed. In both models, the input drive \textbf{in} connection weight is pre-scaled by $\tau$. If a neuron is active, it fires every time-step with a graded spike equal to the soft-threshold activation. As the V1 activation is sparse across the layer, the total number of spikes per time-step remains small. 

\subsection{Residual Neurons}
In 2-layer LCA, the residual neurons are not inherently sparse and must be modified as such for performance. Accumulator neurons \cite{accumulator_neurons} previously have been demonstrated as a way to create a spiking residual layer \cite{dictionarylearning}. In our implementation we expand on accumulator neurons to take advantage of the graded spikes on Loihi 2. The residual neurons accumulate the reconstruction error $e_{t+1} = e_{t} + \textbf{X} - \textrm{in}$ until they reach some error threshold $\lambda_e$. Then the neuron fires a graded spike with magnitude $e$ and resets $e$ to zero. By increasing the error threshold until multiple time steps are required to produce a spike, firing activity can be made more sparse in time. This sparsity dramatically increases throughput and typically doesn't impact the algorithm's solution quality. We explore the impact of the error threshold further in Section \ref{section:two-layer}.

\subsection{Connections}
All connections weights are fixed point values with 8-bit mantissa and an exponent that is constant across the connection. As a result, the precision of the weights is dependent on the maximum magnitude connection. In the two-layer model, connections correspond to unit length feature vectors whose elements don't vary in magnitude sufficiently to be truncated.  The lateral connections in the 1-layer model, whose elements are given by the inner product of feature vectors, can vary much more in comparison. Thus, in the one-layer model weak excitation and inhibition between neurons may be lost with the limited precision. In our later experiments we only observe this behavior at relatively low V1 thresholds.

\section{Benchmarking Setup}

\begin{figure}
\centering
\includegraphics[width=0.9\linewidth]{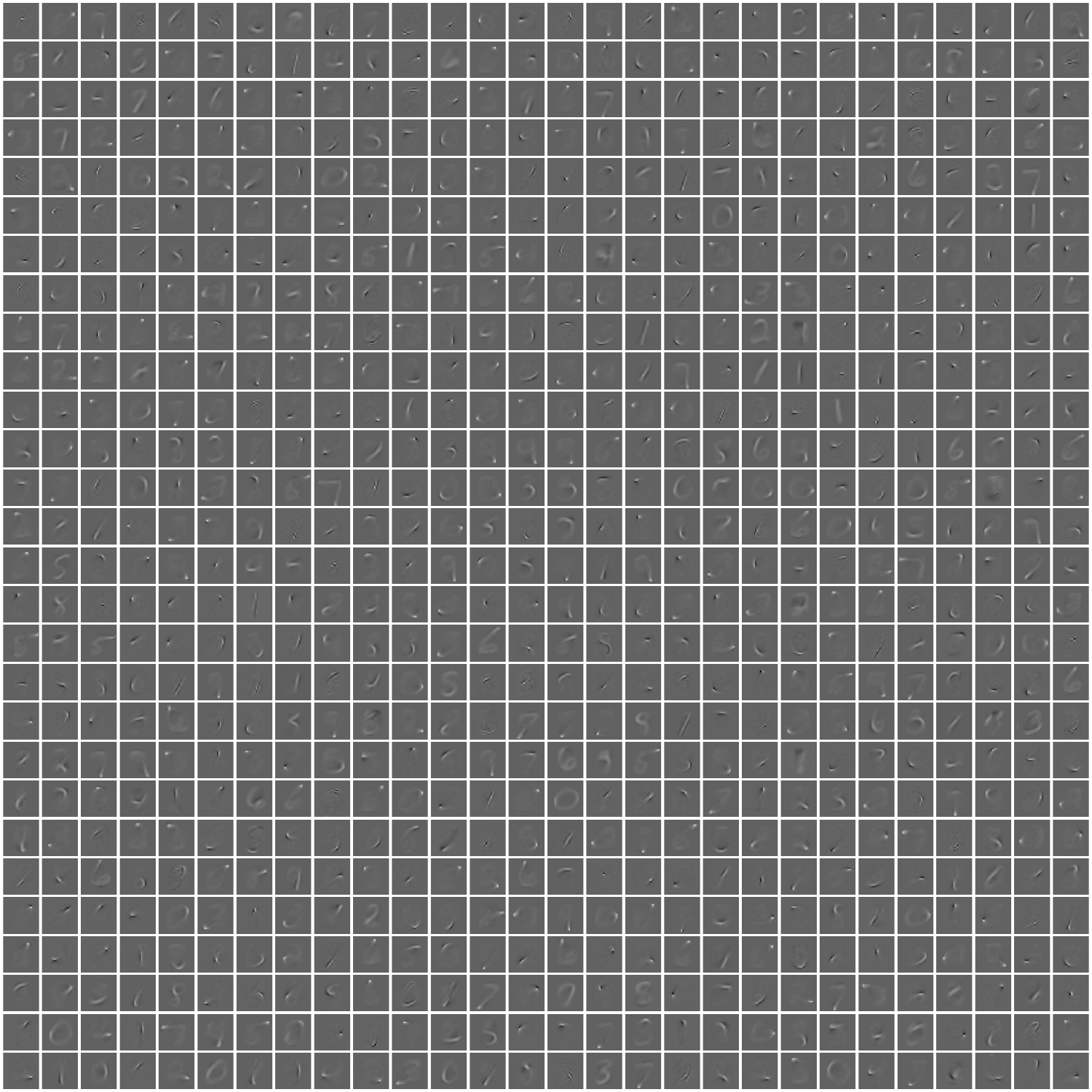}
\caption{Pre-trained dictionary used for benchmarking}
\label{fig:dictionary}
\end{figure}

\begin{table}
    \caption{Devices and libraries used for benchmarking}
    \label{tab:devices}
    \centering
    \begin{tabular}{cccc}
        \toprule
        Platform & Type & Process & Framework (Ver.)\\
        \midrule
        Loihi 2 & Loihi & Intel 4 & Lava (w/ Loihi ext. 0.4.1) \\
        M1 Pro & CPU & TSMC N5 & Accelerate (Xcode 14.2)\\
        Jetson Nano & GPU & TSMC 20nm & Jax \cite{jax2018github} (0.2.25)\\
        A100 & GPU & TSMC N7 & Jax \cite{jax2018github} (0.3.21) \\
        \bottomrule
    \end{tabular}
\end{table}

To evaluate the performance of our implementation, we benchmark the reconstruction of 28px $\times$ 28px grayscale images from MNIST \cite{lecun2010mnist}. Reconstructions use a complete (784 element), pre-trained dictionary shown in Figure \ref{fig:dictionary}. The dictionary is pre-trained on the MNIST training set using SPORCO \cite{brendt_wohlberg-proc-scipy-2017}. We compare LCA on Loihi 2 with other LCA implementations for the CPU and GPU devices listed in Table \ref{tab:devices}. LCA is run for 256 iterations with a time constant of $\tau=2^{-7}$ and one input image at a time (i.e., batch-size = 1 regime). This emulates frame-by-frame sparse-coding of a real-time input. Note that LCA will not fully converge after this many iterations, so performance should not be compared to other sparse coding algorithms. We measure reconstruction error, sparsity, dynamic power, and run-time. On the CPU and GPUs dynamic power is calculated by measuring idle power for 10 minutes, and subtracting it from average power running LCA for 10 minutes. By calculating dynamic power, we mitigate the impact of other running processes on the energy measurements.

\subsection{Loihi 2}

The input to LCA on Loihi 2 is stored in memory that is currently difficult to modify at run-time. To simplify our implementation and limit the impact of IO, we fix the input to LCA for a given run. We instead simulate the effect of changing inputs by resetting the voltage of the V1 neurons every 256 time-steps and averaging results across a small subset of images. All voltages and activations are represented as 24-bit signed fixed point values, with the binary point at $[0000\,0001\,.\,0000\,0000\,0000\,0000]_2$. This gives a range of $[-128_{10}, 127_{10}]$ with precision $2^{-16}$. On the chip the 1-Layer model utilizes 33 cores and the 2-Layer uses 66. Power is measured using Lava's built-in profiler. \footnote{All Loihi results are obtained using Lava extension for Loihi v0.4.1 on {\tt Oheo Gulch} board (system name {\tt ncl-ext-og-01}, available on Intel Neuromorphic Research Community (INRC) cloud \cite{inrcweb}). Results may vary.}

\subsection{CPU and GPU}

We implement 1-Layer LCA on the CPU and GPUs listed in Table \ref{tab:devices}. As the devices can perform a non-local learning rule they don't benefit from 2-Layer LCA, so we do not evaluate it. Both the CPU and GPU implementations use 32-bit floating point values for all computation. We did not find a substantial change in run-time or reconstruction accuracy by using 16-bit values. For GPU implementations we utilize the Jax library and for the M1 Pro we utilize the Accelerate framework. Power is measured using {\tt powermetrics} on MacOS, {\tt nvidia-smi} for the A100, and through onboard sensors on the Jetson Nano. \footnote{CPU and GPU architecture details are as mentioned in Table \ref{tab:devices}.}

\section{Results}
\begin{figure}
    \begin{subfigure}[t]{0.115\textwidth}
        \centering
        \includegraphics[width=0.99\linewidth]{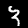}
        \caption{Original}
    \end{subfigure}%
    \hfill
    \begin{subfigure}[t]{0.115\textwidth}
        \centering
        \includegraphics[width=0.99\linewidth]{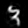}
        \caption{1-Layer\\Loihi 2}
    \end{subfigure}%
    \hfill
    \begin{subfigure}[t]{0.115\textwidth}
        \centering
        \includegraphics[width=0.99\linewidth]{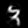}
        \caption{2-Layer\\Loihi 2}
    \end{subfigure}
    \hfill
    \begin{subfigure}[t]{0.115\textwidth}
        \centering
        \includegraphics[width=0.99\linewidth]{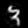}
        \caption{CPU/GPU}
    \end{subfigure}
    
    \caption{Example LCA Reconstructions for Loihi 2 and CPU/GPU Implementations. V1 Threshold $\lambda = 0.5$}
    \Description{TODO describe figure}
    \label{fig:reconstructions}
\end{figure}

\begin{figure}
\centering
\includegraphics[width=\linewidth]{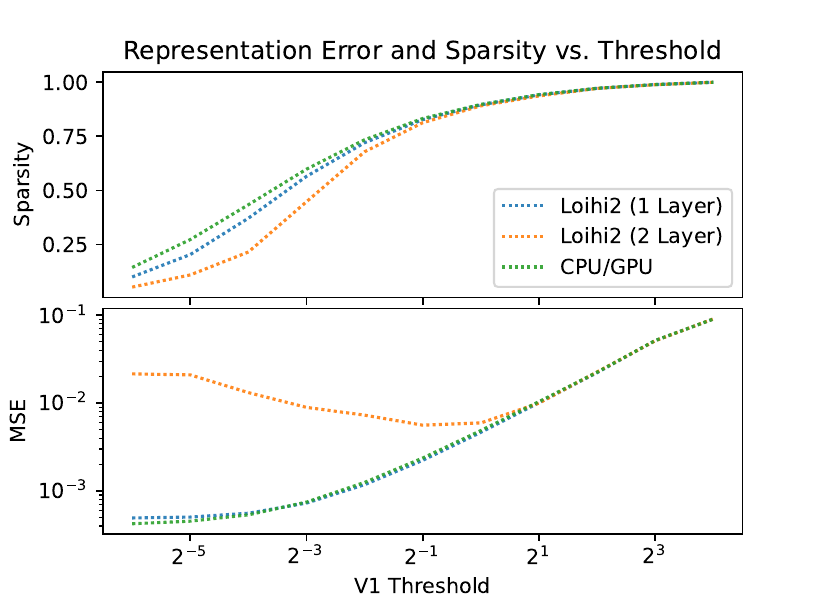}
\caption{Sparse Coding performance across all V1 thresholds. For 2-Layer LCA, $\lambda_e=64$.}
\label{fig:sparsity}
\end{figure}

For 10 randomly sampled images, the reconstruction error and sparsity of solutions obtained using 1-layer LCA on Loihi 2 closely match those of CPU and GPU implementations across most V1 thresholds (Figure \ref{fig:reconstructions}, \ref{fig:sparsity}). At $\lambda < 2^{-6}$ over 90\% of the V1 neurons are active and at $\lambda>2^{4}$ no neurons are active, so we evaluate performance within this range. The reduced weight precision of the 1-layer model relative to CPU/GPU weights results in slightly less sparse solutions at low V1 thresholds. Additionally at these low thresholds the reconstruction error of 2-layer LCA increases notably. It is possible to reduce this error by decreasing the residual threshold, explored in Section \ref{section:two-layer}.

\subsection{Tuning 2-Layer LCA} \label{section:two-layer}

\begin{figure}
\flushright
\includegraphics[width=\linewidth]{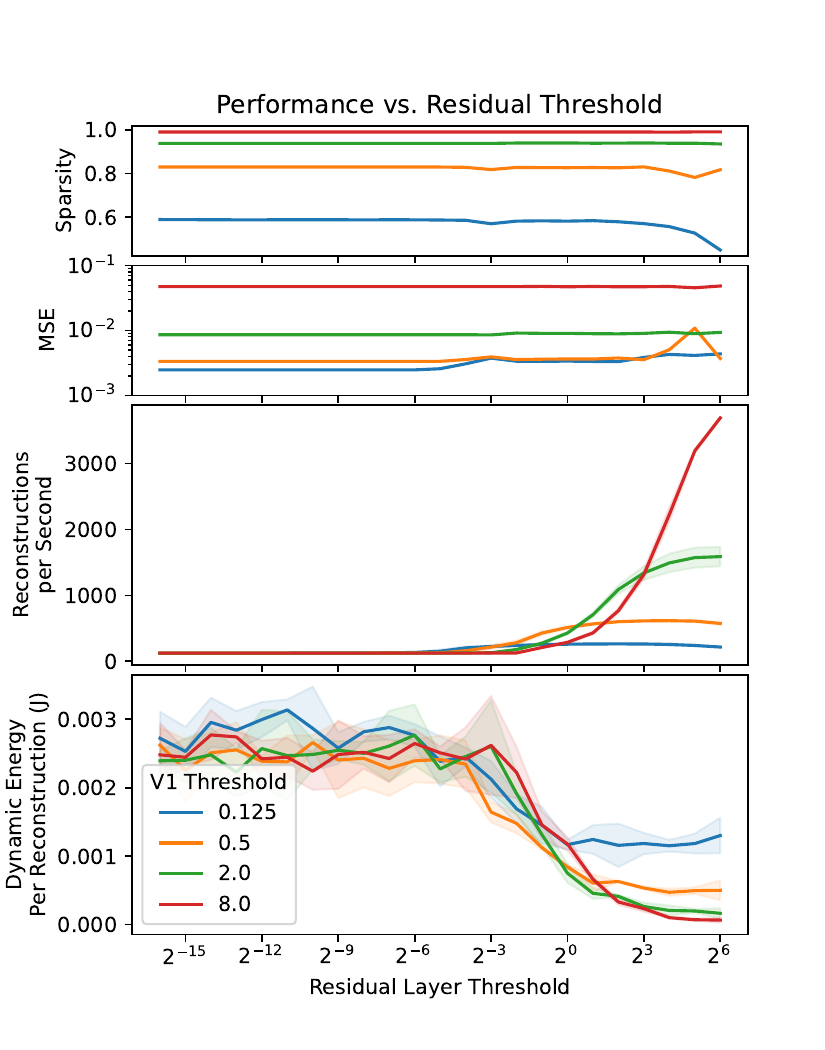}
\caption{Sparse Coding performance across all residual layer thresholds $\lambda_e$. 95\% confidence intervals shown.}
\label{fig:res_threshold}
\end{figure}

It is not immediately clear that LCA converges to similar solutions with a spiking residual layer. To test this, we vary the threshold $\lambda_{e}$ in the residual layer of 2-Layer LCA exponentially from $2^{-16}$ to $2^{6}$ and evaluate the reconstruction of 3 images. Figure \ref{fig:res_threshold} illustrates the impact of different $\lambda_{e}$ values on sparsity, MSE, dynamic, reconstruction throughput, and energy usage. As the threshold increases, the number of spikes decreases and we observe a decrease in both energy consumption and run-time. We find the sparsity and reconstruction error is unchanged for large V1 thresholds, but as the V1 threshold decreases lower residual thresholds must be used for optimal solutions. As there is minimal differences in the solution at larger V1 thresholds but substantially faster and lower energy performance, we utilize $\lambda_{e}=2^6$ for the rest of our tests.

\begin{table*}
    \caption{Overall performance of all devices. $\lambda = 2^{-1}$ for Loihi 2, see Appendix \ref{apdx:tables} for full results. Batch size of 1 unless otherwise noted. $\lambda_e=2^6$ for 2-Layer LCA. }
    \begin{tabular}{lrrrrrr}
        \toprule
        \multirow{2}{6em}{Device}&  \multicolumn{3}{c}{Power (W)} &  \multirow{2}{7em}{Reconstructions Per Second} &  \multirow{2}{9em}{Dynamic Energy Per Reconstruction (mJ)} & \multirow{2}{7em}{Energy Delay Product (mJ $\cdot$ ms)} \\
        & Static & Dynamic & Total & & \\
        \midrule
        Loihi 2 (1 Layer) & 0.54 & 0.39 & 0.93 & 621 & 0.63 & 1.02\\
        Loihi 2 (2 Layer) & 0.55 & 0.30 & 0.84 & 564 & 0.53 & 0.93\\
        M1 Pro & 0.05 & 5.35 & 5.40 & 185 & 28.87 & 155.78\\
        Jetson Nano & 0.13 & 3.48 & 3.61 & 17 & 205.52 & 12,148.68\\
        A100 & 49.83 & 55.52 & 105.35 & 464 & 119.77 & 258.38\\
        \midrule
        A100 (Batch size $2^{14}$) & 52.34 & 319.29 & 371.63 & 140,749 & 2.27 & 4.90\\
        \bottomrule
    \end{tabular}
\end{table*}

\subsection{Performance vs. Threshold}

\begin{figure}
\flushright
\includegraphics[width=\linewidth]{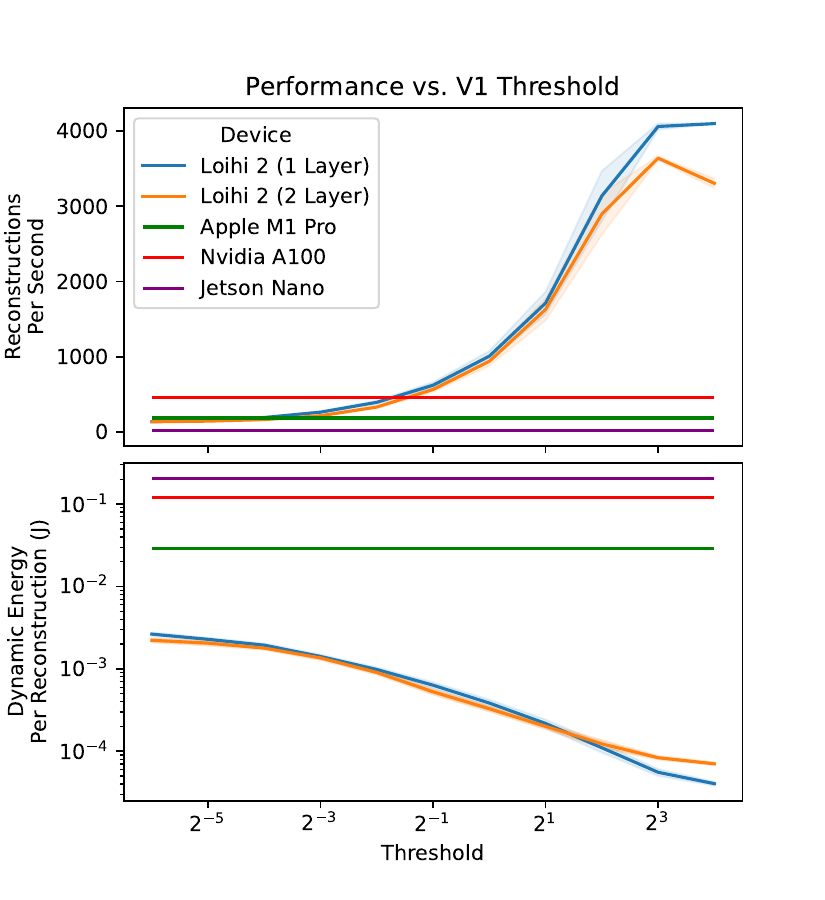}
\caption{Sparse Coding performance across all V1 thresholds $\lambda$. 95\% confidence intervals shown.}
\label{fig:v1_threshold}
\end{figure}

As the runtime and energy usage of Loihi 2 are dependent on the number of spikes per time-step, we evaluate the reconstruction of 10 images for V1 thresholds from $2^{-6}$ to $2^4$.  Figure \ref{fig:v1_threshold} depicts how the runtime and energy usage vary substantially based on the chosen threshold. At V1 thresholds $\lambda \ge 2^{-1}$ 1-Layer LCA on Loihi 2 is faster and more efficient than LCA on all other devices. This threshold corresponds to an average sparsity of 83\%. The 2-Layer implementation is very close in performance across all thresholds. Energy improvements as the threshold increases are driven by increased throughput, but at high thresholds power also decreases.  It is important to note that while the larger threshold values run substantially faster on Loihi 2, the reconstruction quality is reduced on all platforms (Figure \ref{fig:sparsity}). 

With a less optimal dictionary, we observed LCA was unstable on all platforms when the V1 threshold was small. This instability is expected for small $\lambda$ and can be mitigated by reducing $\tau$ \cite{rozell2008sparse}. In these unstable cases LCA performs similarly to the minimum V1 threshold; The Loihi 2 implementation remains power efficient but is substantially slower than both the A100 and M1 Pro.

\section{Conclusion and Future Work}

The benchmarking results of our LCA implementation on Loihi 2 indicate significant advantages in power efficiency and run-time to other hardware platforms. With a V1 threshold $\lambda \ge 2^{-1}$, the solutions obtained on Loihi 2 are roughly identical to those on other devices and are obtained at least 34\% faster with an order of magnitude improvement in power efficiency. 
This study enables the development of intelligent, autonomous, real-time data processing systems on small robots, drones, and satellites, where there are strict SWaP requirements. By demonstrating the superior performance of LCA on Loihi 2 compared to conventional computing devices, this study provides a foundation for future research and development of neuromorphic computing systems, paving the way for a new era of power-efficient, intelligent computing.
Further improvements in speed and efficiency are possible with the Loihi 2 implementation by increasing the V1 threshold $\lambda$, but involve trade-offs in reconstruction quality. Additionally, while slower than the 1-Layer version, 2-Layer LCA is similarly fast and efficient across most thresholds on Loihi 2. This may enable efficient dictionary learning on the platform in the future. As our current experiments only involve up to 10 images being reconstructed with a fixed size dictionary, future work will investigate how these differences change as the dictionary size scales up and compare to other CPU and GPU algorithms. Additionally, it may be possible to improve upon the V1 neuron model to make it spike less frequently using sigma-delta or accumulator dynamics.

\begin{acks}
This material is based upon work supported by the Department of Energy, Office of Science, Advanced Scientific Computing Research program, under award number 77902.
\end{acks}
 
\bibliographystyle{ACM-Reference-Format}
\bibliography{citations}


\begin{thebibliography}{17}

%%% the Web address.

\ifx \showCODEN    \undefined \def \showCODEN     #1{\unskip}     \fi
\ifx \showDOI      \undefined \def \showDOI       #1{#1}\fi
\ifx \showISBNx    \undefined \def \showISBNx     #1{\unskip}     \fi
\ifx \showISBNxiii \undefined \def \showISBNxiii  #1{\unskip}     \fi
\ifx \showISSN     \undefined \def \showISSN      #1{\unskip}     \fi
\ifx \showLCCN     \undefined \def \showLCCN      #1{\unskip}     \fi
\ifx \shownote     \undefined \def \shownote      #1{#1}          \fi
\ifx \showarticletitle \undefined \def \showarticletitle #1{#1}   \fi
\ifx \showURL      \undefined \def \showURL       {\relax}        \fi
% invisible to TeX
\providecommand\bibfield[2]{#2}
\providecommand\bibinfo[2]{#2}
\providecommand\natexlab[1]{#1}
\providecommand\showeprint[2][]{arXiv:#2}

\bibitem[Bradbury et~al\mbox{.}(2018)]%
        {jax2018github}
\bibfield{author}{\bibinfo{person}{James Bradbury}, \bibinfo{person}{Roy
  Frostig}, \bibinfo{person}{Peter Hawkins}, \bibinfo{person}{Matthew~James
  Johnson}, \bibinfo{person}{Chris Leary}, \bibinfo{person}{Dougal Maclaurin},
  \bibinfo{person}{George Necula}, \bibinfo{person}{Adam Paszke},
  \bibinfo{person}{Jake Vander{P}las}, \bibinfo{person}{Skye
  Wanderman-{M}ilne}, {and} \bibinfo{person}{Qiao Zhang}.}
  \bibinfo{year}{2018}\natexlab{}.
\newblock \bibinfo{booktitle}{\emph{{JAX}: composable transformations of
  {P}ython+{N}um{P}y programs}}.
\newblock Google.
\newblock
\urldef\tempurl%
\url{http://github.com/google/jax}
\showURL{%
\tempurl}


\bibitem[{B}rendt {W}ohlberg(2017)]%
        {brendt_wohlberg-proc-scipy-2017}
\bibfield{author}{\bibinfo{person}{{B}rendt {W}ohlberg}.}
  \bibinfo{year}{2017}\natexlab{}.
\newblock \showarticletitle{{S}{P}{O}{R}{C}{O}: {A} {P}ython package for
  standard and convolutional sparse representations}. In
  \bibinfo{booktitle}{\emph{{P}roceedings of the 16th {P}ython in {S}cience
  {C}onference}}, \bibfield{editor}{\bibinfo{person}{{K}aty {H}uff},
  \bibinfo{person}{{D}avid {L}ippa}, \bibinfo{person}{{D}illon {N}iederhut},
  {and} \bibinfo{person}{{M} {P}acer}} (Eds.). \bibinfo{pages}{1 -- 8}.
\newblock
\urldef\tempurl%
\url{https://doi.org/10.25080/shinma-7f4c6e7-001}
\showDOI{\tempurl}


\bibitem[Carroll et~al\mbox{.}(2017)]%
        {Carroll_denoising}
\bibfield{author}{\bibinfo{person}{Jacob Carroll}, \bibinfo{person}{Nils
  Carlson}, {and} \bibinfo{person}{Garrett~T. Kenyon}.}
  \bibinfo{year}{2017}\natexlab{}.
\newblock \showarticletitle{Phase Transitions in Image Denoising via Sparsely
  Coding Convolutional Neural Networks}.
\newblock \bibinfo{journal}{\emph{CoRR}}  \bibinfo{volume}{abs/1710.09875}
  (\bibinfo{year}{2017}), \bibinfo{numpages}{4}~pages.
\newblock
\showeprint[arXiv]{1710.09875}
\urldef\tempurl%
\url{http://arxiv.org/abs/1710.09875}
\showURL{%
\tempurl}


\bibitem[Davies et~al\mbox{.}(2021)]%
        {loihi1}
\bibfield{author}{\bibinfo{person}{Mike Davies}, \bibinfo{person}{Andreas
  Wild}, \bibinfo{person}{Garrick Orchard}, \bibinfo{person}{Yulia
  Sandamirskaya}, \bibinfo{person}{Gabriel A.~Fonseca Guerra},
  \bibinfo{person}{Prasad Joshi}, \bibinfo{person}{Philipp Plank}, {and}
  \bibinfo{person}{Sumedh~R. Risbud}.} \bibinfo{year}{2021}\natexlab{}.
\newblock \showarticletitle{Advancing Neuromorphic Computing With Loihi: A
  Survey of Results and Outlook}.
\newblock \bibinfo{journal}{\emph{Proc. IEEE}} \bibinfo{volume}{109},
  \bibinfo{number}{5} (\bibinfo{year}{2021}), \bibinfo{pages}{911--934}.
\newblock
\urldef\tempurl%
\url{https://doi.org/10.1109/JPROC.2021.3067593}
\showDOI{\tempurl}


\bibitem[Henke et~al\mbox{.}(2022)]%
        {appletospikes}
\bibfield{author}{\bibinfo{person}{Kyle Henke}, \bibinfo{person}{Michael Teti},
  \bibinfo{person}{Garrett Kenyon}, \bibinfo{person}{Ben Migliori}, {and}
  \bibinfo{person}{Gerd Kunde}.} \bibinfo{year}{2022}\natexlab{}.
\newblock \showarticletitle{Apples-to-Spikes: The First Detailed Comparison of
  LASSO Solutions Generated by a Spiking Neuromorphic Processor}. In
  \bibinfo{booktitle}{\emph{Proceedings of the International Conference on
  Neuromorphic Systems 2022}} (Knoxville, TN, USA)
  \emph{(\bibinfo{series}{ICONS '22})}. \bibinfo{publisher}{Association for
  Computing Machinery}, \bibinfo{address}{New York, NY, USA}, Article
  \bibinfo{articleno}{19}, \bibinfo{numpages}{8}~pages.
\newblock
\showISBNx{9781450397896}
\urldef\tempurl%
\url{https://doi.org/10.1145/3546790.3546811}
\showDOI{\tempurl}


\bibitem[Intel(2018)]%
        {inrcweb}
Intel \bibinfo{year}{2018}\natexlab{}.
\newblock \bibinfo{booktitle}{\emph{Intel Neuromorphic Research Community}}.
\newblock Intel.
\newblock
\urldef\tempurl%
\url{https://www.intel.com/content/www/us/en/research/neuromorphic-community.html}
\showURL{%
\tempurl}


\bibitem[LeCun et~al\mbox{.}(2010)]%
        {lecun2010mnist}
\bibfield{author}{\bibinfo{person}{Yann LeCun}, \bibinfo{person}{Corinna
  Cortes}, {and} \bibinfo{person}{CJ Burges}.} \bibinfo{year}{2010}\natexlab{}.
\newblock \bibinfo{booktitle}{\emph{MNIST handwritten digit database}}.
\newblock ATT Labs.
\newblock
\urldef\tempurl%
\url{http://yann.lecun.com/exdb/mnist}
\showURL{%
\tempurl}


\bibitem[Olshausen and Field(1997)]%
        {sparse_V1}
\bibfield{author}{\bibinfo{person}{Bruno~A. Olshausen} {and}
  \bibinfo{person}{David~J. Field}.} \bibinfo{year}{1997}\natexlab{}.
\newblock \showarticletitle{Sparse coding with an overcomplete basis set: A
  strategy employed by V1?}
\newblock \bibinfo{journal}{\emph{Vision Research}} \bibinfo{volume}{37},
  \bibinfo{number}{23} (\bibinfo{year}{1997}), \bibinfo{pages}{3311--3325}.
\newblock
\showISSN{0042-6989}
\urldef\tempurl%
\url{https://doi.org/10.1016/S0042-6989(97)00169-7}
\showDOI{\tempurl}


\bibitem[Orchard et~al\mbox{.}(2021)]%
        {loihi2}
\bibfield{author}{\bibinfo{person}{Garrick Orchard},
  \bibinfo{person}{Edward~Paxon Frady}, \bibinfo{person}{Daniel Ben~Dayan
  Rubin}, \bibinfo{person}{Sophia Sanborn}, \bibinfo{person}{Sumit~Bam
  Shrestha}, \bibinfo{person}{Friedrich~T. Sommer}, {and} \bibinfo{person}{Mike
  Davies}.} \bibinfo{year}{2021}\natexlab{}.
\newblock \showarticletitle{Efficient Neuromorphic Signal Processing with Loihi
  2}.
\newblock \bibinfo{journal}{\emph{CoRR}}  \bibinfo{volume}{abs/2111.03746}
  (\bibinfo{year}{2021}), \bibinfo{numpages}{6}~pages.
\newblock
\showeprint[arXiv]{2111.03746}
\urldef\tempurl%
\url{https://arxiv.org/abs/2111.03746}
\showURL{%
\tempurl}


\bibitem[Parpart et~al\mbox{.}(2022)]%
        {dictionarylearning}
\bibfield{author}{\bibinfo{person}{Gavin Parpart}, \bibinfo{person}{Carlos
  Gonzalez~Rivera}, \bibinfo{person}{Terrence Stewart}, \bibinfo{person}{Edward
  Kim}, \bibinfo{person}{Jocelyn Rego}, \bibinfo{person}{Andrew O'Brien},
  \bibinfo{person}{Steven Nesbit}, \bibinfo{person}{Garrett Kenyon}, {and}
  \bibinfo{person}{Yijing Watkins}.} \bibinfo{year}{2022}\natexlab{}.
\newblock \showarticletitle{Dictionary Learning with Accumulator Neurons}. In
  \bibinfo{booktitle}{\emph{Proceedings of the International Conference on
  Neuromorphic Systems 2022}} (Knoxville, TN, USA)
  \emph{(\bibinfo{series}{ICONS '22})}. \bibinfo{publisher}{Association for
  Computing Machinery}, \bibinfo{address}{New York, NY, USA}, Article
  \bibinfo{articleno}{11}, \bibinfo{numpages}{9}~pages.
\newblock
\showISBNx{9781450397896}
\urldef\tempurl%
\url{https://doi.org/10.1145/3546790.3546801}
\showDOI{\tempurl}


\bibitem[Rozell et~al\mbox{.}(2008)]%
        {rozell2008sparse}
\bibfield{author}{\bibinfo{person}{Christopher~J. Rozell},
  \bibinfo{person}{Don~H. Johnson}, \bibinfo{person}{Richard~G. Baraniuk},
  {and} \bibinfo{person}{Bruno~A. Olshausen}.} \bibinfo{year}{2008}\natexlab{}.
\newblock \showarticletitle{{Sparse Coding via Thresholding and Local
  Competition in Neural Circuits}}.
\newblock \bibinfo{journal}{\emph{Neural Computation}} \bibinfo{volume}{20},
  \bibinfo{number}{10} (\bibinfo{date}{Oct.} \bibinfo{year}{2008}),
  \bibinfo{pages}{2526--2563}.
\newblock
\showISSN{0899-7667}
\urldef\tempurl%
\url{https://doi.org/10.1162/neco.2008.03-07-486}
\showDOI{\tempurl}


\bibitem[Schultz et~al\mbox{.}(2014)]%
        {schultz2014replicating}
\bibfield{author}{\bibinfo{person}{Peter~F. Schultz}, \bibinfo{person}{Dylan~M.
  Paiton}, \bibinfo{person}{Wei Lu}, {and} \bibinfo{person}{Garrett~T.
  Kenyon}.} \bibinfo{year}{2014}\natexlab{}.
\newblock \showarticletitle{Replicating Kernels with a Short Stride Allows
  Sparse Reconstructions with Fewer Independent Kernels}.
\newblock \bibinfo{journal}{\emph{CoRR}}  \bibinfo{volume}{abs/1406.4205}
  (\bibinfo{year}{2014}), \bibinfo{numpages}{12}~pages.
\newblock
\showeprint[arxiv]{1406.4205}~[q-bio.QM]


\bibitem[Teti et~al\mbox{.}(2022)]%
        {TetiKMM22}
\bibfield{author}{\bibinfo{person}{Michael Teti}, \bibinfo{person}{Garrett
  Kenyon}, \bibinfo{person}{Ben Migliori}, {and} \bibinfo{person}{Juston
  Moore}.} \bibinfo{year}{2022}\natexlab{}.
\newblock \showarticletitle{{LCAN}ets: Lateral Competition Improves Robustness
  Against Corruption and Attack}. In \bibinfo{booktitle}{\emph{Proceedings of
  the 39th International Conference on Machine Learning}}
  \emph{(\bibinfo{series}{Proceedings of Machine Learning Research},
  Vol.~\bibinfo{volume}{162})}, \bibfield{editor}{\bibinfo{person}{Kamalika
  Chaudhuri}, \bibinfo{person}{Stefanie Jegelka}, \bibinfo{person}{Le~Song},
  \bibinfo{person}{Csaba Szepesvari}, \bibinfo{person}{Gang Niu}, {and}
  \bibinfo{person}{Sivan Sabato}} (Eds.). \bibinfo{publisher}{PMLR},
  \bibinfo{address}{Baltimore, MD, USA}, \bibinfo{pages}{21232--21252}.
\newblock
\urldef\tempurl%
\url{https://proceedings.mlr.press/v162/teti22a.html}
\showURL{%
\tempurl}


\bibitem[Voelker et~al\mbox{.}(2020)]%
        {accumulator_neurons}
\bibfield{author}{\bibinfo{person}{Aaron~R. Voelker}, \bibinfo{person}{Daniel
  Rasmussen}, {and} \bibinfo{person}{Chris Eliasmith}.}
  \bibinfo{year}{2020}\natexlab{}.
\newblock \showarticletitle{A Spike in Performance: Training Hybrid-Spiking
  Neural Networks with Quantized Activation Functions}.
\newblock \bibinfo{journal}{\emph{CoRR}}  \bibinfo{volume}{abs/2002.03553}
  (\bibinfo{year}{2020}), \bibinfo{numpages}{18}~pages.
\newblock
\showeprint[arXiv]{2002.03553}
\urldef\tempurl%
\url{https://arxiv.org/abs/2002.03553}
\showURL{%
\tempurl}


\bibitem[Watkins et~al\mbox{.}(2018a)]%
        {LCA_compression}
\bibfield{author}{\bibinfo{person}{Yijing Watkins}, \bibinfo{person}{Oleksandr
  Iaroshenko}, \bibinfo{person}{Mohammad Sayeh}, {and} \bibinfo{person}{Garrett
  Kenyon}.} \bibinfo{year}{2018}\natexlab{a}.
\newblock \showarticletitle{Image Compression: Sparse Coding vs. Bottleneck
  Autoencoders}. In \bibinfo{booktitle}{\emph{2018 IEEE Southwest Symposium on
  Image Analysis and Interpretation (SSIAI)}}. \bibinfo{publisher}{IEEE},
  \bibinfo{address}{Las Vegas, NV, USA}, \bibinfo{pages}{17--20}.
\newblock
\urldef\tempurl%
\url{https://doi.org/10.1109/SSIAI.2018.8470336}
\showDOI{\tempurl}


\bibitem[Watkins et~al\mbox{.}(2018b)]%
        {LCA_event_recons}
\bibfield{author}{\bibinfo{person}{Yijing Watkins}, \bibinfo{person}{Austin
  Thresher}, \bibinfo{person}{David Mascarenas}, {and}
  \bibinfo{person}{Garrett~T. Kenyon}.} \bibinfo{year}{2018}\natexlab{b}.
\newblock \showarticletitle{Sparse Coding Enables the Reconstruction of
  High-Fidelity Images and Video from Retinal Spike Trains}. In
  \bibinfo{booktitle}{\emph{Proceedings of the International Conference on
  Neuromorphic Systems}} (Knoxville, TN, USA) \emph{(\bibinfo{series}{ICONS
  '18})}. \bibinfo{publisher}{Association for Computing Machinery},
  \bibinfo{address}{New York, NY, USA}, Article \bibinfo{articleno}{8},
  \bibinfo{numpages}{5}~pages.
\newblock
\showISBNx{9781450365444}
\urldef\tempurl%
\url{https://doi.org/10.1145/3229884.3229892}
\showDOI{\tempurl}


\bibitem[Zhu and Rozell(2013)]%
        {zhu2013visual}
\bibfield{author}{\bibinfo{person}{Mengchen Zhu} {and}
  \bibinfo{person}{Christopher~J Rozell}.} \bibinfo{year}{2013}\natexlab{}.
\newblock \showarticletitle{Visual nonclassical receptive field effects emerge
  from sparse coding in a dynamical system}.
\newblock \bibinfo{journal}{\emph{PLoS computational biology}}
  \bibinfo{volume}{9}, \bibinfo{number}{8} (\bibinfo{year}{2013}),
  \bibinfo{pages}{e1003191}.
\newblock


\end{thebibliography}

\appendix
\begin{table*}[b!]
\section{Appendix} \label{apdx:tables}

    \caption{1-Layer Model Results on Loihi 2 for all $\lambda$}
    \begin{tabular}{lcccrrr}
        \toprule
        \multirow{2}{1em}{$\lambda$}&  \multicolumn{3}{c}{Power (W)} &  \multirow{2}{7em}{Reconstructions Per Second} &  \multirow{2}{9em}{Dynamic Energy Per Reconstruction (mJ)} & \multirow{2}{7em}{Energy Delay Product (mJ $\cdot$ ms)} \\
        & Static & Dynamic & Total & & \\
        \midrule
        $2^{-6}$ & 0.54 & 0.36 & 0.91 & 138 & 2.64 & 19.14 \\
        $2^{-5}$ & 0.54 & 0.35 & 0.90 & 156 & 2.28 & 14.68 \\
        $2^{-4}$ & 0.54 & 0.37 & 0.91 & 192 & 1.94 & 10.09 \\
        $2^{-3}$ & 0.54 & 0.37 & 0.91 & 263 & 1.41 & 5.37 \\
        $2^{-2}$ & 0.54 & 0.39 & 0.93 & 394 & 0.98 & 2.49 \\
        $2^{-1}$ & 0.54 & 0.39 & 0.93 & 621 & 0.63 & 1.02 \\
        $2^{0}$ & 0.54 & 0.38 & 0.93 & 1,008 & 0.38 & 0.38 \\
        $2^{1}$ & 0.54 & 0.37 & 0.91 & 1,713 & 0.22 & 0.13 \\
        $2^{2}$ & 0.54 & 0.34 & 0.88 & 3,136 & 0.11 & 0.04 \\
        $2^{3}$ & 0.54 & 0.23 & 0.77 & 4,058 & 0.06 & 0.01 \\
        $2^{4}$ & 0.54 & 0.17 & 0.71 & 4,096 & 0.04 & 0.01 \\    
        \bottomrule
    \end{tabular}
\end{table*}

\begin{table*}
    \caption{2-Layer Model Results on Loihi 2 for all $\lambda$}
    \begin{tabular}{lcccrrr}
        \toprule
        \multirow{2}{1em}{$\lambda$}&  \multicolumn{3}{c}{Power (W)} &  \multirow{2}{7em}{Reconstructions Per Second} &  \multirow{2}{9em}{Dynamic Energy Per Reconstruction (mJ)} & \multirow{2}{7em}{Energy Delay Product (mJ $\cdot$ ms)} \\
        & Static & Dynamic & Total & & \\
        \midrule
        $2^{-6}$ & 0.54 & 0.30 & 0.84 & 135 & 2.22 & 16.50 \\
        $2^{-5}$ & 0.54 & 0.30 & 0.84 & 144 & 2.05 & 14.29 \\
        $2^{-4}$ & 0.54 & 0.29 & 0.84 & 164 & 1.79 & 10.87 \\
        $2^{-3}$ & 0.54 & 0.29 & 0.84 & 216 & 1.36 & 6.30 \\
        $2^{-2}$ & 0.54 & 0.30 & 0.84 & 331 & 0.90 & 2.73 \\
        $2^{-1}$ & 0.54 & 0.30 & 0.84 & 564 & 0.53 & 0.93 \\
        $2^{0}$ & 0.54 & 0.31 & 0.85 & 939 & 0.33 & 0.35 \\
        $2^{1}$ & 0.54 & 0.32 & 0.87 & 1,629 & 0.20 & 0.12 \\
        $2^{2}$ & 0.54 & 0.35 & 0.89 & 2,895 & 0.12 & 0.04 \\
        $2^{3}$ & 0.54 & 0.30 & 0.85 & 3,638 & 0.08 & 0.02 \\
        $2^{4}$ & 0.54 & 0.23 & 0.78 & 3,303 & 0.07 & 0.02 \\
        \bottomrule
    \end{tabular}
\end{table*}

% \begin{table*}
%     \begin{tabular}{lrrrrr}
%     \caption{Overall performance of all devices. Loihi 2 results are averaged over all stable V1 thresholds. $\lambda_e=2^6 for $2-Layer LCA}
%     \toprule
%     \multirow{2}{6em}{Device}&  \multicolumn{3}{|c|}{Power (W)} &  \multirow{2}{7em}{Reconstructions Per Second} &  \multirow{2}{9em}{Dynamic Energy Per Reconstruction (mJ)}\\
%     & Static & Dynamic & Total & & \\
%     \midrule
%     Loihi 2 (1 Layer) & 0.54 & 0.25 & 0.79 & 2498 & 0.21 \\
%     Loihi 2 (2 Layer) & 0.51 & 0.33 & 0.84 & 2255 & 0.32 \\
%     M1 Pro & 0.05 & 5.35 & 5.40 & 185 & 28.87 \\
%     Jetson Nano & 0.13 & 3.48 & 3.61 & 17 & 119.77 \\
%     A100 & 49.83 & 55.52 & 105.35 & 464 & 205.52 \\
%     \bottomrule
%     \end{tabular}
% \end{table*}

\end{document}